# Ontologies for the Integration of Air Quality Models and 3D City Models

Claudine Metral

Institut d'architecture - University of Geneva

Site de Battelle - 7, route de Drize - CH 1227 Carouge/Geneva - Switzerland

claudine.metral@unige.ch

Gilles Falquet

Department of Information Systems - University of Geneva

Site de Battelle - 7, route de Drize - CH 1227 Carouge/Geneva - Switzerland

gilles.falquet@cui.unige.ch

Kostas Karatzas,

Informatics Applications and Systems Group, Dept. of Mechanical Engineering

Aristotle University, GR-54124 Thessaloniki, Greece

kkara@eng.auth.gr

## *Abstract*

In the perspective of a sustainable urban planning, it is necessary to investigate cities in a holistic way and to accept surprises in the response of urban environments to a particular set of strategies. For example, the process of inner-city densification may limit air pollution, carbon emissions, and energy use through reduced transportation; on the other hand, the resulting street canyons could lead to local levels of pollution that could be higher than in a low-density urban setting.

The holistic approach to sustainable urban planning implies using different models in an integrated way that is capable of simulating the urban system. As the interconnection of such models is not a trivial task, one of the key elements that may be applied is the description of the urban geometric properties in an "interoperable" way. Focusing on air quality as one of the most pronounced urban problems, the geometric aspects of a city may be described by objects such as those defined in CityGML, so that an appropriate air quality model can be applied for estimating the quality of the urban air on the basis of atmospheric flow and chemistry equations.

It is generally admitted that an ontology-based approach can provide a generic and robust way to interconnect different models. However, a direct approach, that consists in establishing correspondences between concepts, is not sufficient in the present situation. One has to take into account, among other things, the computations involved in the correspondences between concepts.

In this paper we first present theoretical background and motivations for the interconnection of 3D city models and other models related to sustainable development and urban planning. Then we present a practical experiment based on the interconnection of CityGML with an air quality model. Our approach is based on the creation of an ontology of air quality models and on the extension of an ontology of urban planning process (OUPP) that acts as an ontology mediator.

# 1. Introduction

**Sustainable development and urban planning**

One of the most often cited definitions of sustainability is the one created by the Brundtland Commission in its report (Brundtland, 1987) which defined sustainable development as development that "meets the needs of the present without compromising the ability of future generations to meet their own needs." The report highlighted three fundamental components to sustainable development: environmental protection, economic growth and social equity.

Regardless of whether environmental issues are not the only aspects of sustainable planning and decision processes, they are impossible to circumvent. In urban areas, one of the most important environmental problems is air pollution, mostly induced by vehicle traffic. This pollution can cause severe damages on health and is particularly crucial because of the high density of population in cities. The improvement of air quality is therefore imperative and must be taken into account in a sustainable urban planning process. In fact a deep understanding and an accurate prediction of urban flow and pollutant dispersion in cities is required for a more proactive urban planning. In addition, the existing EU legal framework (Dir 96/62 and the associated "daughter" directives), dictate the use of air quality models as one of the ways to be applied for atmospheric quality assessment and related policy making.

**Air quality models**

Air quality models are important tools to study, understand and predict air pollution levels. Among the various categories, 3D air pollution models are those that aim at "reconstructing" the environment, its properties and governing physical laws. Current air quality models and the associated simulation systems apply a number of (simplifying) assumptions about the environment, in particular its geometry and its physical properties, according to the scale of interest. Thus, models that are used for the study of air quality at a local-to-regional scale make use of a grid-base spatial resolution, where each grid cell has dimensions in the order of magnitude of 1 x 1 km. Coming down to the urban scale, simulation models try to reconstruct the building shapes and basic geometrical characteristics and dimensions, in order to improve simulation results.

**3D city models**

At the same time, 3D city models are rapidly developing (a number of cities have already been modeled in CityGML). These models could provide the air quality models with (some of) the missing information they require to be more precise and effective. On the other hand, urban planners and stakeholders need information about the impact of their development plans on air quality evolution (current air quality and future scenario projections). Thus their models must be complemented with data originating from air quality models. However, these data do not fit directly into the urban planning models, they must be adapted (aggregated, computed, etc.) to these models.

3## 2. CityGML

**What is CityGML?**

CityGML is an open information model for the representation and exchange of virtual 3D City Models on an international level (OGC, 2006) (Kolbe et al, 2005) (Kolbe et al, 2006). CityGML defines the classes and relations for the most relevant topographic objects in cities and regional models with respect to their geometrical, topological, semantical, and appearance properties. It has been defined in this way because purely graphical or geometrical 3D models, if appropriate for visualization purposes, are not sufficient for applications such as urban and landscape planning, architectural design, touristic and leisure activities, 3D cadastres, environmental simulations, mobile telecommunications, disaster management, homeland security, vehicle and pedestrian navigation, training simulators or mobile robotics.

CityGML differentiates five consecutive Levels of Detail (LOD), from terrain alone to architectural models (outside and interior). With LOD increasing, objects become more detailed, both at the geometry and the thematic level. Besides their attributes, objects in CityGML can have external references to corresponding objects in other databases or data sets. For example, a building in CityGML may be related to a cadastral database.

With the Application Domain Extensions (ADE), CityGML provides an extension mechanism so as to enrich the data with identifiable features for specific domain areas, such as environmental noise mapping.

CityGML is implemented as an XML application schema of the Geography Markup Language 3 (GML3), the extendible international standard for spatial data exchange and encoding issued by the Open Geospatial Consortium (OGC) and the ISO TC211.

**Underlying concepts of CityGML**

The main thematic fields are:
- the terrain (named as Relief Feature),
- the coverage by land use objects (named as Land Use),
- transportation (both graph structures and 3D surface data),
- vegetation (solitary objects, areas and volumes, with vegetation classification),
- water objects (volumes and surfaces),
- sites, in particular buildings (bridge, tunnel, excavation or embankment in the future),
- City Furniture (for fixed object such as traffic lights, traffic signs, benches or bus stops).

So the main underlying concepts of CityGML (see figure below) are :
Relief Feature, Land Use, Site, Building (as subconcept of Site), Transportation Object, Vegetation Object, Water Object, City Furniture.
Buildings in CityGML can be represented with roofs, walls, windows, doors, rooms and even furniture (depending on the LOD) associated to the concepts Roof Surface, Wall Surface, Window, Door, Room and Building Furniture (see figure below).

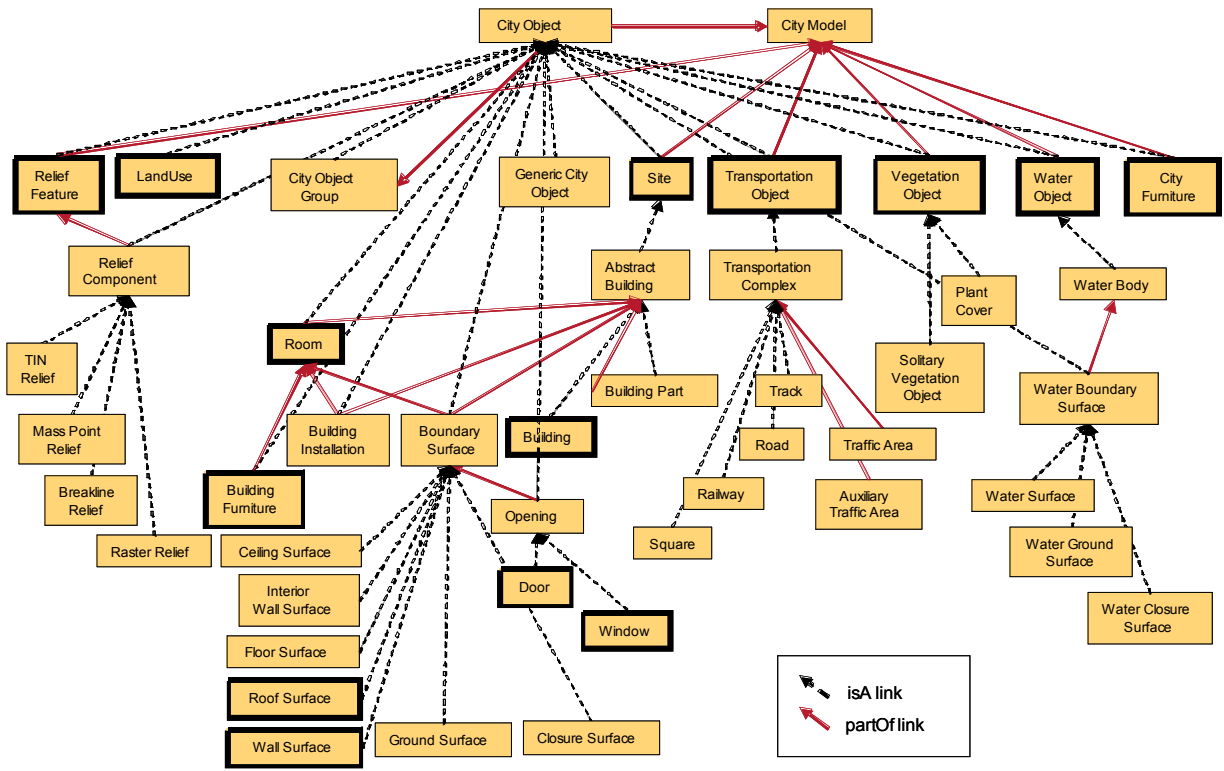

## 3. OUPP: an ontology of urban planning

CityGML, with its urban and geometric concepts and their semantic relationships, is useful for representing an urban planning project, but not sufficient. Indeed, some non geometric concepts such as right of way or more abstract concepts such as soft mobility don't exist in CityGML. Similarly, relationships such as is used by or participates to are missing (see figure below). This is why we decided to define an ontology of urban planning process (OUPP).

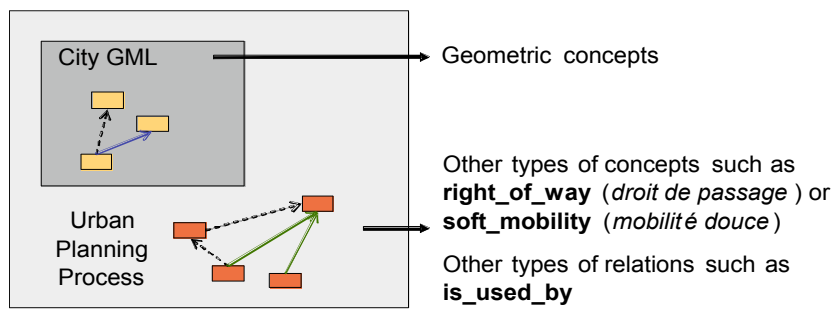

OUPP has been defined and is still under development in the framework of the "Integrating Urban Knowledge into 3D City Models" research project funded by the Swiss Confederation and part of the European COST Action "Towntology". The main objective of this research is to contribute to a better communication between the various actors involved in an urban planning process. The main sources for OUPP are master and local plans (such as PLQ), as well as GIS data of Geneva.

An urban planning project (that uses urban data and documents such as spatial data from GIS, 3D city models, texts, maps, plans, pictures) is represented by an ontology with concepts and instances related to the data and documents. With this representation, a user can not only view in an integrated way the project but also accede only to the knowledge that fits its profile and centers of interest (Metral et al, 2006).

In our research, use of CityGML is done from OUPP. More precisely, OUPP refers to the concepts that exist in CityGML. An ontology alignment has been realized between OUPP and CityGML, with 1 to 1, 1 to many and many to 1 alignment relations (an example of a n to 1 alignment relation is given in figure below).

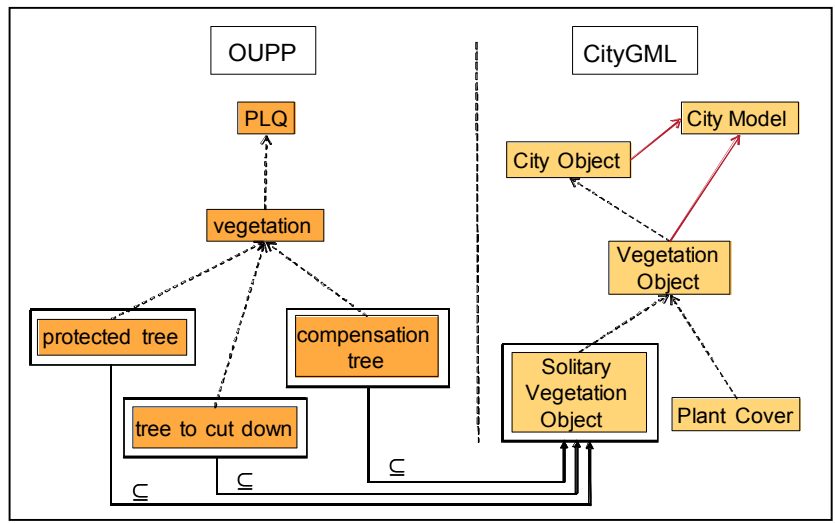

# 4. Air quality models

**The aims of air quality models**

In urban areas, traffic-related pollution is one of the more environmental crucial problems (EEA, 2006). As air pollution entails damages on humans, vegetation and ecosystems, reducing pollutant emissions to the atmosphere is imperative. Models of air pollutant dispersion, transport and transformation have been defined. They are used for supporting such reductions and, more generally, for supporting environmental decision making

**The diversity of air quality models**

According to (Moussiopoulos et al 2006), models describing the dispersion and transport of air pollutants in the atmosphere can be distinguished in many ways such as:
- the spatial scale (global; regional-to-continental; local-to-regional; local),
- the temporal scale (episodic models, (statistical) long-term models),
- the treatment of the transport equations (Eulerian, Lagrangian models),
- the treatment of various processes (chemistry, wet and dry deposition)
- the complexity of the approach.

**Urban scale models**

At this scale, in general, air flow is very complex, as it depends not only on the meteorological conditions but also on the characteristics of urban obstacles such as buildings (form, orientation with regard to the wind direction, etc.). The air flow patterns that develop around buildings govern the dispersion mechanisms and dictate concentration levels in a built environment.

Air quality may become seriously poor in densified urban areas. Street canyons, that are relatively narrow streets between buildings lining up continuously along both sides, are more and more frequent in such areas. Unfortunately they can induce serious pollution problems because pollutants emitted in a street canyon (such as vehicle exhaust gases) tend to disperse less than those emitted in an open area, while the combination of pollutants emitted and local meteorological conditions may trigger pollution production mechanisms that result in accumulated concentration levels and bad air quality

**Street canyon models**

Various types of numerical models have been employed to simulate flow and dispersion of pollutants in urban street canyons. While most are two-dimensional models such as (Baik & Kim, 1999) (Huang et al, 2000), there exists some three-dimensional models such as (Kim and Baik, 2004) (Santiago et al, 2007).

The flow and pollutant dispersion in urban street canyons are mainly controlled by meteorological conditions (such as the prevailing wind direction and speed) and canyon geometry, in particular the street aspect ratio (ratio of the building height to the width between buildings). When wind blows over buildings with different shapes and sizes, it is disturbed by the buildings, and turbulent eddies, that can stream into the street canyon, are produced. The turbulence intensity, which depends on the geometric configuration of the surrounding buildings, has an effect on the pollutant dispersion in the street canyon (Kim & Baik, 2003). Furthermore, when the street bottom or the building wall is heated by solar radiation, thermally induced flow is formed that combines with mechanically induced flow formed in the canyon (Kim & Baik, 2001) (Huang et al, 2002) (Liu et al, 2003). Lastly, pollutants emitted from automobiles, for example NO and NO2, are chemically reactive (Baik et al, 2007).

Air pollution modeling within street canyons has to deal with problems at the so called neighborhood scale, i.e. 100 m to 1 km. As the flow around a building is influenced by the geometry and the scale of the obstacle (Fig 1), the modeling attempts applied so far focus on the representation of both aspects of the urban web via detailed 3d representations. For this reason, mesh-based 3D modeling is applied, trying to represent as much as possible the actual geometry, scale, and physical characteristics of the area of interest (Fig 2).

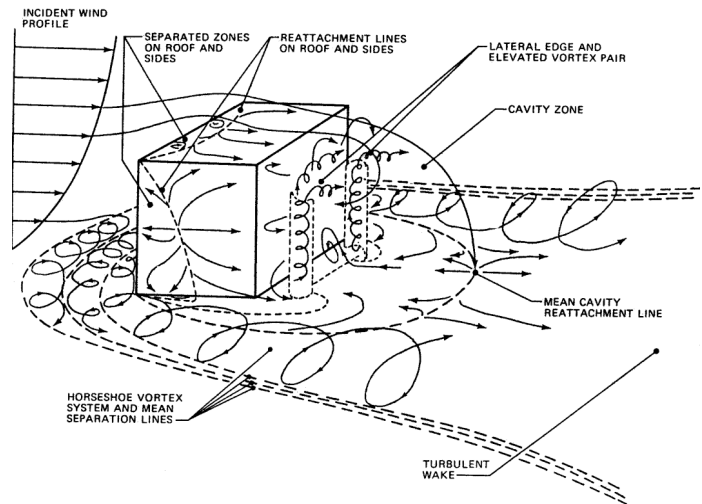

Figure 1: Flow around a building (cube)

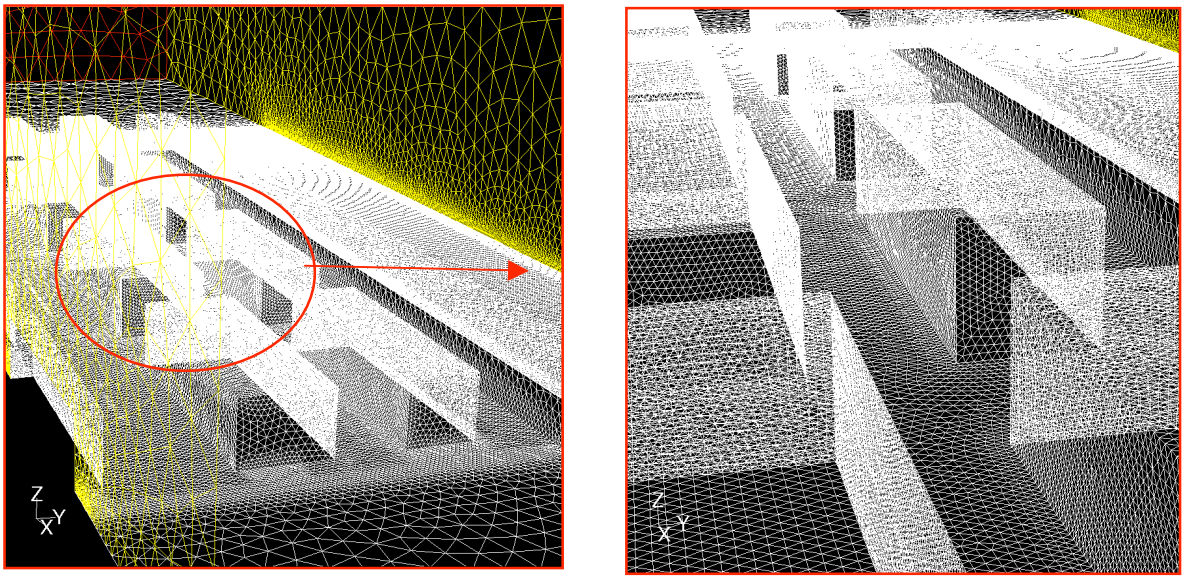

Figure 2: 3-d modeling within a city. Figure taken from E. Solazzo: Advanced Tools for Rational Energy Use towards Sustainability with emphasis on microclimatic issues in urban applications. Presentation made duting the ATREUS project meeting in Thessaloniki,Greece
(http://aix.meng.auth.gr/atreus/thessaloniki_meeting.html)

The basic problem resulting from this approach is the fact that it is very demanding when it comes to computational resources. In addition to that, the mesh-based simulation needs to be supplemented by the proper parameters, representing the thermal characteristics of the surfaces, their relative position in accordance to the sun, mechanically generated turbulence (from traffic). These types of problems could be automatically solved, if urban modeling were possible via the use of semantically rich 3D objects (such as CityGML objects). For instance, the thermal characteristics of a surface could be derived from data such as surface geometry and orientation, material, color, etc. that exist in 3D city models. In addition, some of the most important parameters like the roughness length in urban airflow modeling, also result "automatically" from 3D data about obstacle shape, dimensions and relative position, thus making the use of a 3D urban model extremely advantageous in comparison with the way that it is currently being done.

Studies realized with street canyon models revealed that favorable meteorological conditions and appropriate urban geometry can reduce pollutant concentration levels (Kovar-Panskus et al., 2002). Therefore, results generated by these models are crucial for a sustainable and proactive urban planning process. On the other hand, street canyon models take as input parameters relative to the geometry, the topology and the semantics of streets and buildings. Such data are available in 3D city models expressed in CityGML, either directly, either in another form and possibly through some computations. Thus, the interconnection of air quality models with CityGML and OUPP is a first step towards taking into account the city in a more holistic way.

## 5. Towards an ontology of air quality models

The first phase of our ontology of air quality models focuses on street canyon models.

**Underlying concepts and properties of street canyon models**

Street canyon models are based on equations taking as input:
- the **pollutant source** characteristics (source location, emitted product, etc.)
- the **meteorological conditions**, mainly the ambient wind conditions (speed, direction relative to the street canyon, etc.) but also, to some extent, the thermal conditions (solar heating)
- the **street canyon** geometry**,** in particular its aspect ratios (height-to-width ratio for 2D and 3D models, height-to-length ratio for 3D models) or its orientation relative to the ambient wind

and producing as output:
- a **flow** characterized mainly by its **vortices** (intensity, rotation direction, location, etc.)
- a **pollutant dispersion distribution**.

Thus, the main underlying concepts are: Street Canyon, Meteorological Conditions, Pollutant Source, Flow , Vortex (as part of Flow) and Pollutant Dispersion Distribution. The concepts Pollutant Source and Pollutant Dispersion Distribution refer to each other. The Ambient Wind Conditions and Thermal Conditions are also main concepts, defined as part of Meteorological Conditions. As a street canyon is the space between buildings that line up continuously along both sides of a relatively narrow street (Liu et al, 2003), the concept Street Canyon has to be defined from the concepts Street and Building. A flow and its vortices are both vector fields. Thus the concept Vector Field has to be added as superconcept of Flow and of Vortex. The pollutant concentration distribution is a scalar field that we want to associate to isosurfaces for visualization of the values. Thus we also need the concepts Scalar Field and Isosurface (see figure below).

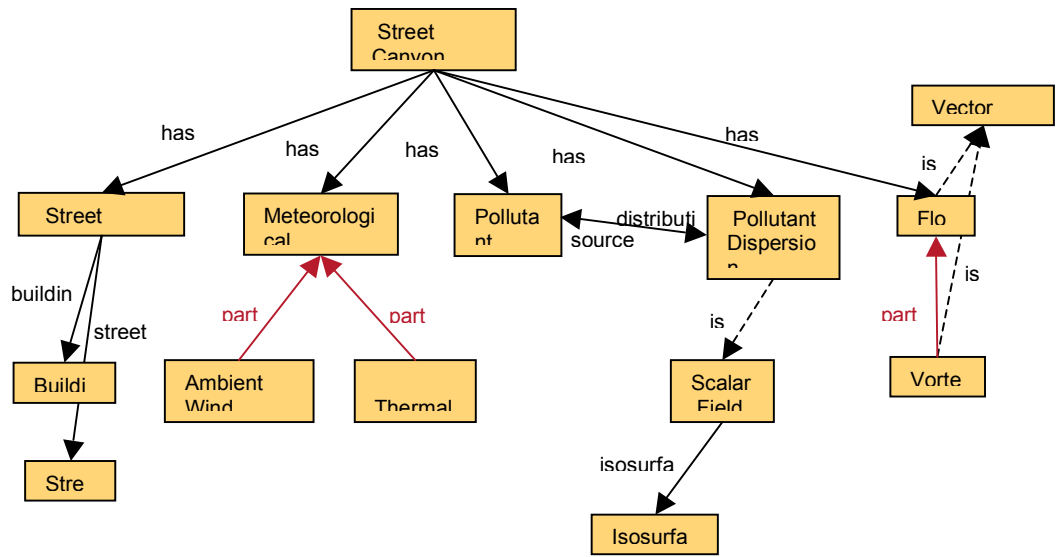

Figure 3: The Street Canyon part of the air quality ontology

The concepts with their properties are shown below:

**Street Canyon**
    shape                        wide / square / narrow
    height-to-width ratio    ratio of the Building height to the Street width
    height-to-height ratio   ratio of the Building height at windward side to the Building height
                                            at leeward side
    orientation               (absolute)
    leeward side location    (depends on the current wind conditions)
    windward side location
    level                           street-level / building-level / roof-level

**Street**
    location                  (absolute)
    width
    orientation              (absolute)
    albedo                   (average reflection coefficient)

**Building**
    location                  (absolute)
    height
    roof slope
    albedo                   (average reflection coefficient)

**Meteorological Conditions**

**Ambient Wind Conditions** (as part of Meteorological Conditions)
    speed
    direction                (absolute)
    turbulence intensity

**Thermal Conditions** (as part of Meteorological Conditions)
   sunshine duration
   temperature of air
   temperature at street bottom
   temperature on surface of building at leeward side
   temperature on surface of building at windward side

The direction of the ambient wind to the street canyon (required in the equations of the model) can be computed from the absolute direction of the ambient wind and the absolute orientation of the street

**Pollutant Source**
   emitted product
   origin                 traffic-related / chemical toxics / biological toxics
   reactivity           reactive / non-reactive
   source location     street-level / advected
   emission rate      continuous / non-continuous
   dispersion distribution  which is a Pollutant Dispersion Distribution

**Flow** (which is a Vector Field)
   regime              isolated roughness flow / wake interference flow / skimming flow

**Vortex**               is a Vector Field
   part of             a Flow
   intensity           weak / strong
   rotation direction    clockwise / counter-clockwise
   location
   origin                mechanically induced / thermally induced
   shape               portal vortex / roll-type vortex / horseshoe vortex

**Isosurface**
   value range
   geometry           a closed surface

**Vector Field**

**Scalar Field**
   isosurface          property which refers to Isosurface concept

**Pollutant Dispersion Distribution** (which is a Scalar Field)
   source              a Pollutant Source

## 6. Interconnecting CityGML and air quality models

In this section we will present different integration patterns that occur when interconnecting CityGML (CG) and an air quality model (AQ) and how to represent them in the OUPP ontology. Our goal is to provide a formal (logical) description of the interconnections at the conceptual level that can easily be turned into an automated (computerized) process. The main idea is to create "associative" concepts and axioms in OUPP that refer to the interconnected concepts in AQ and CG. The aim is
1. to represent potentially complex integration patterns
2. to put all the integration related entities in OUPP and avoid modifying AQ or CG
3. to provide a kind of centralized mediator for integrating other models (water quality, noise reduction, etc.)

The interconnection technique we propose here is inspired by the work of Mitra, Wiederhold and Kersten (2000) on articulation ontologies. In their interconnection architecture the semantic relationships between two source ontologies are described by articulation rules (implications between concepts of the two ontologies), which are then translated to yield concepts in an articulation ontology and semantic implication edges between the articulation ontology and the source ontologies, They also propose functional rules that are intended to normalize values expressed in different systems of measurement. In the following, we extend this approach by defining interconnection concepts that may have different types of semantic links with the source (CG and AQ) ontologies. Another aspect that distinguishes our approach from most of the other approaches (see (Kalfoglou and Schorlemmer, 2003) or (Shvaiko and Eurenat, 2005) for surveys) is the role of the OUPP ontology. In our context, the OUPP is not simply an articulation or a semantic bridge, it is a ontology with its own concepts that may not exist in the other ones.

## *Interconnection patterns*

*0. Property ranges*

The simples interconnection pattern occurs when a property of a concept in one ontology has its range in the other ontology. For instance, the geometry property of an isosurface in AQ is a surface, which is a concept defined in CG. In this situation it is not necessary to define an interconnection concept.

*1. Concepts with different viewpoints*

The properties of the two concepts are not the same in both ontologies and some properties can be computed from those in the other ontology. For instance, street in AQ has a width property while street in CG has a geometry. In addition, the width in AQ can be derived from the geometry in CG.
    We introduce the following interconnection concept (concepts with the same name are distinguished by prefixing them with the ontology name)

**OUPP:street**
    properties
        in_AQ       a AQ:street
        in_CG       a CG:street

Each instance of OUPP:street refers to the same street in AQ and CG.

The following axiom expresses the dependency between the width and geometry properties

**for all** s **in** OUPP:street : s.in_AQ.width = width_computation( s.in_CG.geometry)

where width_computation represents the (possibly complex) geometric function that computes the width of the polygon that forms the street geometry in CG.

Another similar examples is AQ:building, with its roof slope computed from the roof surface of CG:building and its albedo obtained from the surfaces, material, and color of CG:building.

*2. Instances derived from several instances in the other ontology*

This is a more complex integration pattern where a concept instance in one ontology corresponds to a set of concept instances in the other one. For example, a street canyon (in AQ) exists only if there is some street (in CG) bordered by buildings in a particular configuration. The interconnection concept has the following definition:

**OUPP:street_canyon**
    properties
        in_AQ        a AQ:street_canyon
        street        a CG:street
        buildings_1    a set of CG:building
        buildings_2    a set of CG:building

where buildings_1 and buildings_2 refer to the sets of buildings that border the street on both sides. The following axiom defines the notion of street canyon

**for all** c **in** OUPP:street_canyon :
    **for all** x **in** c.buildings_1 : borders(x, s)
    and
    **for all** y **in** c.buildings_2 : borders(y, s)
    and
    continuously_aligned(c.buildings_1)
    and
    continuously_aligned(c.buildings_2)

where borders and continuously_aligned are geometric predicates. In addition, the properties of the street canyon in AQ depend on its components (in CG) properties (as in case 1) and possibly on other factors. For example

**for all** c **in** OUPP:street_canyon : c.height-to-height_ratio =
        average_height(c.buildings_1) / average_height(c.buildings_2)

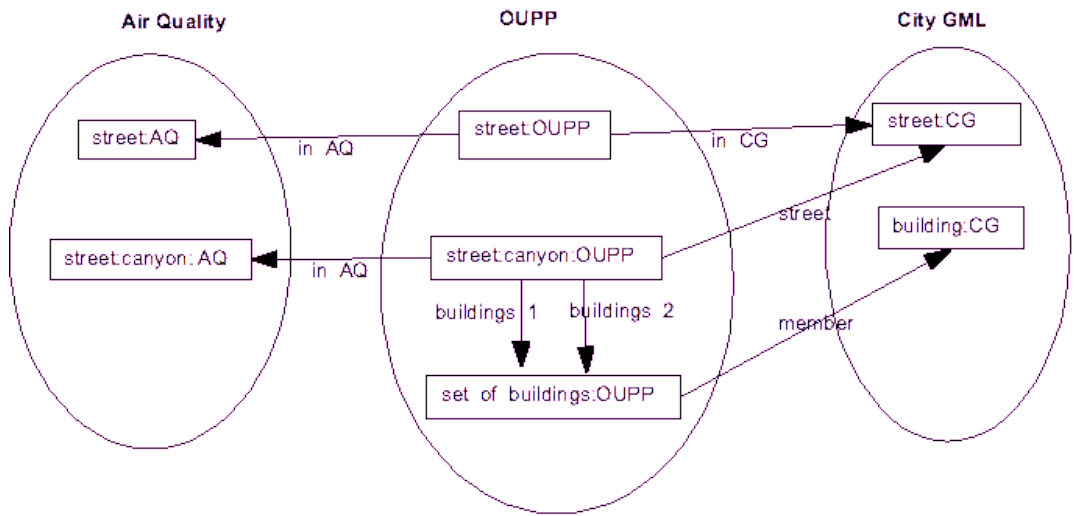

Figure 4: Concepts in OUPP as mediators between the Air Quality and CityGML ontologies

This interconnection technique can be used to "import" theoretical results from one domain into another. For instance, consider the following result:

*An experiment led by Baik and Kim (2002) showed that, with buildings of similar height on both sides of the canyon and a street aspect ratio H/W of 2 (building height of 80m and street width of 40m for example), the pollutant concentration in the lower region of the street canyon is higher near the downwind building than near the upwind building. On the other hand, in the upper region of the canyon, the concentration is higher near the upwind building than near the downwind building.*

As the lower part of the street canyon corresponds to the area most people walk or cycle, it is important to take into account (import) this result in sustainable urban planning. For example, if most of the time the wind blows in the same direction, it is preferable to position a cycle path on the upwind side of the canyon as it is the side with the lower level of pollutant.

This can be formally represented by defining a new concept "street canyon with favourable upwind side" (SCFUS for short) in OUPP. This concept would be associated with the following axiom:

**for all** s **in** OUPP:SCFUS : $(0.9 < x.\text{in\_AQ.height-to-height\_ratio} < 1.1)$
    **and** $(1.9 < \text{height-to-width ratio} < 2.1)$

Similarly, we could define the concept of potentially polluted street as streets that belong to a street canyon where air flow simulation produces a flow with one or more vortices (thus keeping pollutants in the canyon)

From a software engineering point of view, the integration patterns we have shown here specify the computational processes that are necessary to transfer and translate data from one model into the other one.

# 7. Conclusion and future work

In this paper we have argued that the integration of air quality models with 3D city models can greatly improve air quality modeling and sustainable urban planning. Air quality modeling can take advantage of the semantically rich representations of 3D city models to facilitate air flow simulation. The simulation results and other theoretical results from air quality modeling can then enrich 3D city models to help urban planners deal with air pollution issues.

We have proposed a theoretical and practical approach to air quality and 3D city model integration that is based on the definition of an air quality ontology and the use of an urban planning process ontology that acts as a mediator. Through this approach it is possible to support sophisticated interconnection patterns and to formally specify them. In addition, the OUPP mediation ontology can be re-used to integrate several other models that are of interest for sustainable urban planning.

In the near future we plan to enrich this integration example with other results and concepts from air quality modeling. Then we will apply this same approach to integrate other models, in particular water quality models. We will also develop computational tools to effectively re-use and transfer data between models.